\documentclass[10pt,twocolumn,letterpaper]{article}

\usepackage{wacv}
\usepackage{times}
\usepackage{epsfig}
\usepackage{graphicx}
\usepackage{amsmath}
\usepackage{amssymb}

\usepackage{hyperref}       
\usepackage{url}            
\usepackage{booktabs}       
\usepackage{amsfonts}       
\usepackage{nicefrac}       
\usepackage{microtype}      
\usepackage{graphicx}
\usepackage{amsmath}

\usepackage{float}
\newcommand{\bftab}{\fontseries{b}\selectfont}

\newcommand{\todo}[1]{}
\renewcommand{\todo}[1]{{\color{red} TODO: {#1}}}
\usepackage{hyperref}       
\usepackage{url}            
\usepackage{booktabs}       
\usepackage{amsfonts}       
\usepackage{nicefrac}       
\usepackage{microtype}      
\usepackage[T1]{fontenc}

\usepackage{varwidth}
\usepackage{multicol}
\usepackage{booktabs}
\usepackage{multirow}
\usepackage{latexsym}
\usepackage{graphicx}
\usepackage{tabularx}
\usepackage{color,soul}
\usepackage{pbox}
\usepackage{xcolor}

\wacvfinalcopy 

\def\wacvPaperID{778} 

\ifwacvfinal\fi
\begin{document}

\title{Enforcing Reasoning in Visual Commonsense Reasoning}

\author{Hammad A. Ayyubi \hspace{2cm} Md. Mehrab Tanjim \hspace{2cm} David J. Kriegman\\
Department of Computer Science\\
UC San Diego\\
{\tt\small \{hayyubi, mtanjim, kriegman\}@eng.ucsd.edu}
}

\maketitle
\ifwacvfinal\thispagestyle{empty}\fi

\begin{abstract}
    The task of Visual Commonsense Reasoning is extremely challenging in the sense that the model has to not only be able to answer a question given an image, but also be able to learn to reason. The baselines introduced in this task are quite limiting because two networks are trained for predicting answers and rationales separately. Question and image is used as input to train answer prediction network while question, image and correct answer are used as input in the rationale prediction network. As rationale is conditioned on the correct answer, it is based on the assumption that we can solve Visual Question Answering task without any error - which is over ambitious. Moreover, such an approach makes both answer and rationale prediction two completely independent VQA tasks rendering cognition task meaningless. In this paper, we seek to address these issues by proposing an end-to-end trainable model which considers both answers and their reasons jointly. Specifically, we first predict the answer for the question and then use the chosen answer to predict the rationale. However, a trivial design of such a model becomes non-differentiable which makes it difficult to train. We solve this issue by proposing four approaches - softmax, gumbel-softmax, reinforcement learning based sampling and direct cross entropy against all pairs of answers and rationales. We demonstrate through experiments that our model performs competitively against current state-of-the-art. We conclude with an analysis of presented approaches and discuss avenues for further work.
\end{abstract}

\section{Introduction}
In recent years,
computer vision systems have achieved outstanding results
in tasks such as Recognition, Classification, Segmentation and Detection \cite{mahajan2018exploring, chen2019hybrid, liu2018path}. To put the recent successes in perspective, all the aforementioned tasks fall in the category of recognition. Essentially, in most cases the models answer the question ``What?'' or ``Where?'' rather than ``Why''. However, we know that human perception goes well beyond such trivial recognition tasks. By just looking at an image, we are able to deduce many things - contexts, situations, mental states of actors and many more things. Such a higher order of intellect 
is termed as cognition. 

\begin{figure}[t]
    \begin{center}
    \includegraphics[width=0.5\textwidth]{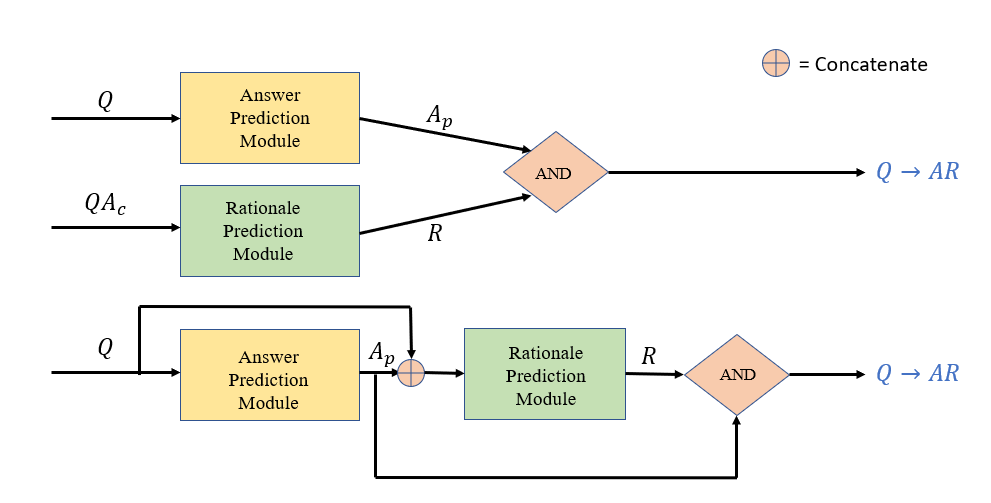}
    \caption{Comparison of our approach against VCR baseline. Top row: baseline approach by Zellers \etal~ \cite{r2c}. Bottom row: our approach. $Q$:Question, $A_c$:Correct Answer, $A_p$: Predicted Answer, $R$: Predicted Rationale, $Q->AR$: Both answer and rationale prediction given question.}
    \end{center}
    \label{comp}
\end{figure}

Cognition is extremely important and relevant.
For example, a higher cognitive ability will help social robots to interact seamlessly with humans. Ability to judge and comprehend mental states of humans will be invaluable to healthcare robots. Additionally, being able to solve this challenging task will help the vision community as a whole to move to the next generation of vision systems which goes beyond normal recognition. 

\begin{figure*}[t]
    \begin{center}
    \includegraphics[width=\textwidth]{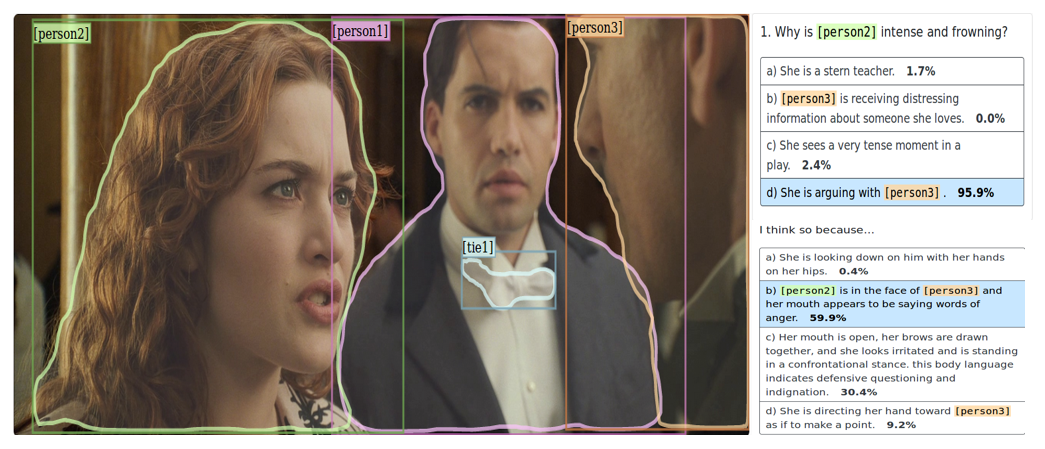}
    \end{center}
    \caption{The Visual Commonsense Reasoning task, Zellers \etal~\cite{r2c}}
    \label{r2c}
\end{figure*}

The main goal of Visual Commonsense Reasoning (VCR) is to solve this cognition task. Precisely, the task is introduced and formulated in 
\cite{r2c} as follows: given an image and a question related to the image, 
the model has to predict the correct answer from four possible choices and 
at the same time, it has to pick the right rationale, again from four options.
As the task is new, Zellers \etal~\cite{r2c} provided a new baseline for it
which seeks to tackle the task of predicting answers and predicting rationales separately. At 
first, answers are predicted given the question and image and then, rationales are predicted given the image and question 
with the correct answer (see figure \ref{comp} and \ref{r2c}). 
This way, the task can be essentially seen as a Visual Question Answering as in both cases the model is trying to predict an answer given the image and query.
Since the rationale prediction module is conditioned on the correct answer while training, the inherent assumption is that the answer prediction network can predict correct answer with 100\% accuracy. This assumption is clearly far fetched as even the state-of-the-art Visual Question Answering (VQA) model can barely reach 75\% accuracy (Kim \etal~\cite{kim2018bilinear}). 
Moreover, as rationale task is carried out independently of the answer prediction task, it is apparent that the model 
fails to capture causal reasoning and the "cognition" ability.

In this paper, we address these issues 
by enforcing the network to consider rationales while predicting the answer (figure \ref{comp}). Specifically, to predict the correct answer for the correct reasons, first we predict the answer given the image and the question. 
Then, using the image, question and the predicted answer, we predict the rationale with the purpose of establishing a bridge for flow of information from rationale prediction module to answer prediction module. However, such an approach will make the end-to-end network non-differentiable because of discrete choices made during training. To solve this problem, we propose four methods: 
\begin{itemize}
    \item Softmax - We first predict the answer probabilities given image and the question. Then, we use the softmax-weighted answers appended to the question as "question" to the rationale module.
    \item Gumbel-Softmax - Similar to our softmax method, except that we use a gumbel-softmax probabilities (Kusner \etal~\cite{gmbl}) to weight our answers to be fed in as question to the rationale module.
    \item Reinforcement Learning Based Sampling - Instead of weighting our answers, we sample an answer according the predicted probability in the first module. We use this sampled answer appended to the question as "question" to the rationale module. We make the end-to-end network differentiable using expectation loss.
    \item Direct Cross Entropy - We train the model to directly predict the correct rationale and answers, given the question and all sixteen pairs of answers and rationales.
\end{itemize}

By adopting these methods, we can get rid of the assumption that answer prediction network has to be 100\% accurate. 
Concurrently, it makes our approaches incomparable to the baselines provided by Zellers \etal~\cite{r2c}. It is so because they always condition their rationale prediction module on correct answer and as such it is bound to be better than a model which conditions on predicted answer.
With this in mind, we propose new baselines in which we train the rationale prediction network by conditioning on correct answer 75\% of the time and random answers for the rest 25\%. Correct answers are provided 75\% of the time to keep a safe estimate of state-of-the-art VQA model. We demonstrate through experiments that our proposed approaches are able to learn the correct answers for the correct reasons. 
Even though our models are not provided with correct answer for rationale prediction, they still perform competitively to the state-of-the-art VCR model. In short, our contributions can be summarised as:

\begin{itemize}
    \item We propose an end-to-end trainable model which considers both answers and their reasons jointly. By doing so, we avoid the unrealistic assumption that for reasoning part the model has to know the correct answer.
    \item To make our model differentiable, we introduce four approaches - softmax, gumbel-softmax, reinforcement learning based sampling and direct cross entropy. This forces the model to predict answers conditioned on the rationale.
    \item We propose a new and proper baseline for the VCR task which feeds correct answer to the rationale prediction module 75\% time.
    \item We experimentally demonstrate that the model learns to predict correct rationales even without being fed with correct answers while still giving comparable performance to current state-of-the-art.
\end{itemize}

\section{Related Work}

The task in Zellers \etal~\cite{r2c} is essentially posed as a question-answering task. Although they have enforced reasoning for the network, the reasoning is still in the question-answer format. As such, it makes sense to explore current work in visual question answer domain. 

A common approach in VQA (Visual Question Answering) is to encode the question and the images into representative vectors, combine the meaning of both vectors using
\cite{an_2, an_3, an_4, an_5, an_6} and use a MLP (Multi-layer Perceptron) with softmax for answer prediction. Agrawal \etal \cite{an_2} use a two layer LSTM \cite{lstm} to encode the question and the last layer of VGGNet \cite{an_8} to encode the image. To normalize image features, l2 norm is used. The image and question features are then fused via element wise multiplication. It is then passed through a fully connected layer followed by a softmax layer to obtain probability distribution over answers. The method, while provided good baseline performance, was naive in its effort to jointly
learn the combined meaning of question and image representation. 

Maaten \etal~\cite{an_3} improved upon the results of \cite{an_2}, by concatenating the question features, image features and response features together, followed by a MLP and softmax. They posed the task as a "yes" or "no" answer by training on question, image and response triplet. This method again naively approached the task of combining question and image features
by only concatenating them. 

Anderson \etal~\cite{an_9} propose an orthogonal work to \cite{r2c}, in which Faster-RCNN \cite{an_10} is used to predict the image regions the model should attend to. We note the difference from our current proposed work - annotations are provided in the VCR 1.0 dataset \cite{r2c} in form of bounding box and segmentation maps.

Akira \etal~\cite{an_4} propose a more sophisticated method to combine the feature vectors from questions and images. After extracting features from questions and images using LSTM and CNN respectively, they use bi-linear pooling (outer vector product) to encode the interplay between image and question representations. This method is much more expressive, but it is also very computationally expensive as outer vector product increases the parameters exponentially. To tackle this, they reduce the full outer vector product to tractable operations using FFT and convolutions. 

While the method was more sophisticated than naive multiplication \cite{an_2} or concatenation \cite{an_3}, it still makes use of some critical assumptions which limits its ability to fully capture the expressive power of outer vector product.

Kim \etal~\cite{an_5} improves upon \cite{an_4} by improving the outer vector product computation using low-rank bi-linear pooling utilizing Hadamard product (elementwise computation). Benyounes \etal~\cite{an_6} further improve upon \cite{an_4} by using Tucker Decomposition of image/question correlation tensor which is able to represent full bi-linear interactions while maintaining the size of the model tractable. 

Zellers \etal~\cite{r2c} propose to jointly learn language and image representation using Bi-LSTM by feeding in image features from CNN for all annotated words. They call this step grounding. Further, query and responses is contextualized using attention mechanism. Finally, the attended query, attended image and response is passed through a Bi-LSTM to make final predictions. One major drawback of the work is that separate networks are trained to predict answers and to reason.

We seek to build on \cite{r2c} by proposing a method to jointly train prediction and reasoning networks. We first choose the correct answer based on the predicted probability distribution using image and question. Then, a combined representation of image, question and chosen answer is fed to reasoning network to select the correct reason. It is to be noted that the step involving choosing an answer to feed to reasoning network is non-differentiable. As such, we propose two ways to tackle this. One, using just a softmax weighted representation of all answers. Two, sampling an answer based on the softmax probability distribution. The network is made differentiable using an expected loss as defined in \cite{rl}.

\begin{figure*}[t]
    \begin{center}
    \includegraphics[width=\textwidth, height=250pt]{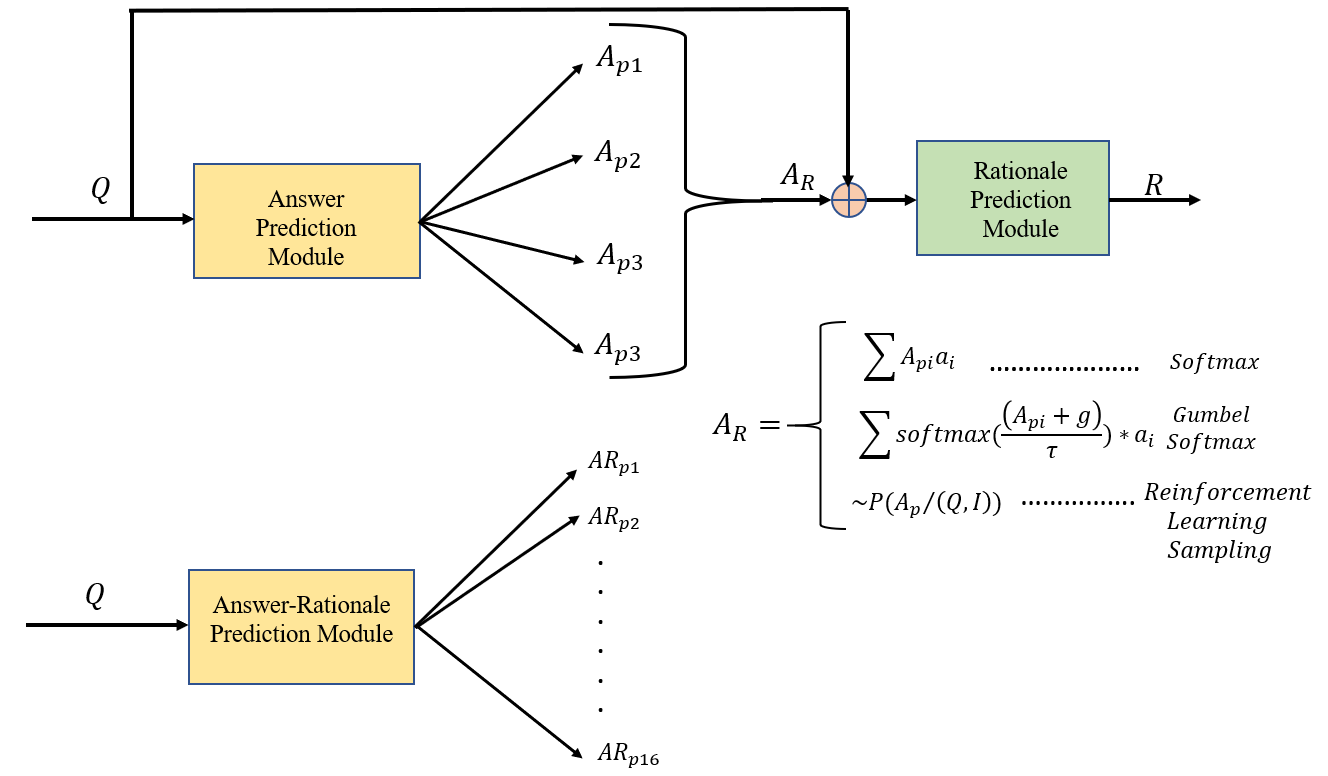}
    \end{center}
    \caption{Our approach. $Q$:Question, $a_i$: $i^th$ answer $A_{pi}$: predicted probability for answer $a_i$, $A_R$: answer representation, $R$: predicted rationale, $AR_{pi}$: predicted probability for answer-rationale combination (4 answer $\times$ 4 rationale = 16 combinations), $I$: image, $\tau$: temperature, $g$: sampled from gumbel distribution.}
    \label{ours}
\end{figure*}

\section{Our Approach}

Given an image and a question we first predict the answer using backbone architecture used by Zellers \etal~\cite{r2c}. We then combine the predicted answer and the question as "question" for the rationale module, and predict the rationale (see figure \ref{ours}). We use the same backbone architecture as in answer prediction part to predict the rationale. We discuss in detail the backbone architecture and each of our four approaches in the following sections.

We use the same backbone architecture to predict answers and rationales as in \cite{r2c}. The questions and response (answers and rationales) are provided as a combination of natural language words and tags for annotated objects in the image. Word embedding for question $\mathit{q}$ and response $\mathit{r}$ are calculated using BERT \cite{bert}. Image features for annotated objects are calculated using ResNet50 \cite{resnet}. The word embedding and image features for tagged objects in the question and response are fed to a bi-directional LSTM \cite{lstm}. This learns a joint language-visual representation vector. \cite{r2c} call this step Grounding.

Next, the response vector is contextualized against the question vector using attention mechanism. In this step an attended question representation is found for every token in the response. Additionally, an attended object representation is found for every response token using similar attention mechanism. This step is called Contextualization.

In the Reasoning step, the joint language-visual representation for response, along with attended question and attended object representation is fed to a bidirectional LSTM. The output of the LSTM is softmaxed to predict the correct answer.

\subsection{Softmax}

In this approach, we first predict the answer probabilities $\mathit{\hat{p}_i}$ using the answer prediction model. We use these probabilities to weight the answers $\mathit{a_i}$. 
\begin{equation}
    A_w = \displaystyle\sum_{i=1}^{4} \hat{p}_i a_i
\end{equation}

This weighted answer answer $A_w$ is appended to the question and fed to the rationale module as query/question. The rationale module considers this (original question and appended weighted answer) as question and the four provided rationales as responses. A model similar to answer prediction part is then used to predict the rationale. 

This approach essentially forms the query for rationale module by appending question and a weighted representation of the answers based on probabilities predicted by the answer prediction module. 

\subsection{Gumbel-Softmax}

In this approach, we use a gumbel-softmax weighted representation of the answer instead of vanilla softmax weighting. Just like in \cite{gmbl}, we use temperature annealing over training period to achieve hard sampling like representation for answers. The gumbel-softmax equation is given by,

\begin{equation}
    \hat{gp} = softmax(1/\tau(\hat{p} + g)
\end{equation}

where $g$ is sampled from gumbel distribution and $\tau$ is the temperature which is annealed over training period. The weights $\hat{gp}$ is then used to weight the answers similar to what we did in the softmax approach. The weighted answer is then appended to the original question and fed to the rationale module as query, exactly similar to how we did previously.

\subsection{Reinforcement Learning Based Sampling}
Prior approaches (softmax and gumbel-softmax) used a weighting of answer representation to feed to the rationale. Of course, this is not same as feeding in the correct (predicted) answer. But, hard choosing makes the network non-differentiable. To tackle this, we use a reinforcement learning based method inspired by \cite{rl}.

First, we sample an answer based on the predicted answer probability distribution $\mathit{\hat{p}}$ (generated by the answer prediction module). This sampled answer is then appended to the question and fed to the rationale prediction module as the query. Rest follow similar to prior approaches.

\begin{equation}
    A \sim P(a/q,I)
\end{equation}

where $P(a/q,I$ is the probability of answer given question and image, which is predicted by the answer prediction module.

The sampling operation, being non-differentiable, is made differentiable using expectation loss from policy gradient approach in reinforcement learning \cite{rl}. It is given by,

\begin{equation}
    loss = \mathbb{E}_{A\sim P(a/q,I)} [\mathit{l}(R/[q,A],I)]
\end{equation}

where $\mathit{l}(R/[q,A],I)$ is the negative log likelihood loss of predicting rationale $R$ given sampled answer $A$ appended to question $q$ and the image, $I$.

\subsection{Direct Cross Entropy}

In this method, we feed as input to network the question and the image. The network is tasked with predicting the right choice from all sixteen combination of answer and rationales. One option out of the sixteen, which contains the correct answer and the correct rationale, is correct. This way the network is forced to consider all sixteen possible combinations and forced to predict the right answer with the right rationale. A cross entropy loss with all the sixteen options is used to train the network.

\begin{figure*}[t]
    \includegraphics[width=\textwidth, height=125pt]{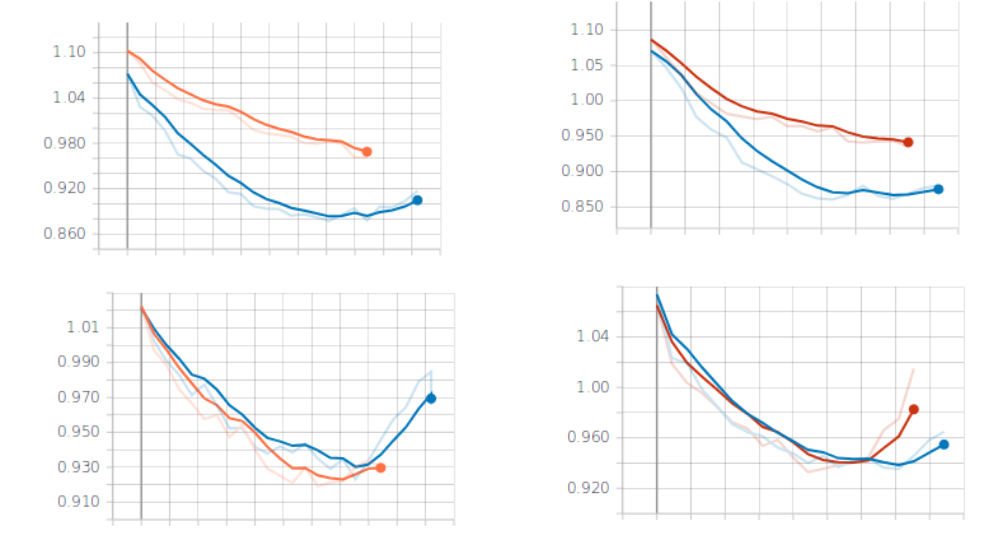}
    \caption{ The left column is softmax model and the right column is gumbel-softmax model. The top row is Q->A loss and the bottom row in QA->R loss. The blue line denotes model trained with Q->A loss : QA->R loss = 1:1, orange/red lines denote model trained with Q->A loss : QA->R loss = 1:4. As can be seen from the curves weighting the QA->R loss four times more results in slight improvement in the QA->R module while significant decline in Q->A module performance.}
    \label{r2c_loss}
\end{figure*}

\section{Experimental Details}

\subsection{Datasets}
The dataset used in all experiments in this work is VCR 1.0 \cite{r2c}. VCR dataset contains 290k multiple choice questions which has been collected from 110k movie scenes. The dataset provides object annotations, labels and classes for all objects in the image. The questions, answers and rationales are quite open ended. A lot of questions seek to ask 'Why?' making the task non-trivial.

\subsection{Experimental Setup}
We use ResNet50 \cite{resnet} as backbone to extract image features in all experiments. The rest of the model has been explained in secion 3.

For training, we use adam optimizer with a learning rate of 2e-4 and weight decay of 1e-4. We use a learning rate strategy which reduces the learning rate by 0.5 every time loss plateaus. We train the model using the whole VCR dataset for 20 epochs as was done in \cite{r2c} to align with the baselines. We also use gradient clipping while training.

We use two losses for all our methods - answer prediction loss and rationale prediction loss, corresponding to each module. All our results have been reported on the validation set of the dataset as test set labels are not available since it's an ongoing challenge. We report test set results only for our best model which was submitted to the leaderboard. We compare only this model with state-of-the-art.

We define certain terms which we are going to use henceforth:
 Q->A - answer prediction network, given question and image,
QA->R - rationale prediction network, given question, image and answer,
 Q->AR - answer and rationale both prediction network, given question and image.

\begin{table}
  \caption{Model performance results (Val set)}
  \centering
  \begin{tabular}{llll}
    \toprule
    Approach     & Q->A     & QA->R     &Q->AR \\
    \midrule
    New baseline     &  63.8\% & 56.4\%      & 38.2\%  \\
    Direct Cross Entropy & 61.55\%  & 13.83\%   & 8.42\%  \\
    RL Sampling     & 57.2\% & 53.2\%      & 34.4\%  \\
    Softmax     & 63.76\% & \textbf{61.61}\%      & 39.76\%  \\
    Gumbel Softmax     & \bftab 64.54\% & 61.05\%      & \textbf{40.28}\%  \\
    \bottomrule
  \end{tabular}
  \label{own_results}
\end{table}

\begin{table}
  \caption{Comparison with the new baseline (Val set)}
  \centering
  \begin{tabular}{llll}
    \toprule
    Approach     & Q->A     & QA->R     &Q->AR \\
    \midrule
    R2C\cite{r2c}     & 63.8\% & \bftab 67.2\%      &  \bftab 43.1\%  \\
    New baseline     &  63.8\% & 56.4\%      & 38.2\%  \\
    Gumbel Softmax     & \bftab 64.54\% & 61.05\%      & 40.28\%  \\
    \bottomrule
  \end{tabular}
  \label{comp_new_baseline}
\end{table}

\begin{table}
  \caption{Comparison with state-of-the-art (Test set)}
  \centering
  \begin{tabular}{llll}
    \toprule
    Approach     & Q->A     & QA->R     &Q->AR \\
    \midrule
    RevisitedVQA\cite{an_3} & 40.5\%  & 33.7\%   & 13.8\%  \\
    BottomUpTopDown\cite{an_9} & 44.1\%  & 25.1\%   & 11.0\%  \\
    MLB\cite{an_5}     & 46.2\% & 36.8\%      & 17.2\%  \\
    MUTAN\cite{an_6}     & 45.5\% & 32.2\%      & 14.6\%  \\
    \cmidrule(r){1-4}
    R2C\cite{r2c}     & 65.1\% & \bftab 67.3\%      &  \bftab 44.0\%  \\
    \cmidrule(r){1-4}
    Gumbel Softmax (ours)     & \bftab 65.7\% & 61.1\%      & 41.1\%  \\
    \bottomrule
  \end{tabular}
  \label{comp_results}
\end{table}

\subsection{Baselines}
As mentioned earlier, our approach is not directly comparable to the baseline provided by Zellers \etal~\cite{r2c} since they feed the correct answer to the rationale module while we feed in the predicted answer. As such, we generate new baseline model.

State-of-the-art VQA model have at max 75\% accuracy, Kim \etal~\cite{kim2018bilinear}. Consequently, it's a reasonable assumption that the answer prediction module can predict the correct answer 75\% time. Keeping this in mind we make our baseline wherein we feed to the rationale module original question appended by the correct answer 75\% time and random answer 25\% time. We leave the answer prediction module as is. 

Finally, we train the two networks separately and combine the results of answer prediction module and rationale prediction module using "AND" operation as was done by Zellers \etal~\cite{r2c}. 

For completeness, we also mention the results of four other baselines from \cite{r2c}. These baseline methods use the ResNet-50 (same as \cite{r2c})  visual architecture and Glove as text representations. These baselines are as follows:

\begin{itemize}
    \item \textbf{RevisitedVQA}\cite{an_3}: This is a version of VQA model which is mainly optimized for response like `yes' and `no'. Basically, it takes a query, response, and image features as inputs and trains by passing the result through MLP layer. 
    \item \textbf{Bottom-up and Top-down attention (BottomUpTopDown)}\cite{an_9}: \cite{r2c} adopted this model as another baseline by passing
object regions referenced by the query and response. The main model attends over region proposals given
by an object detector. 
    \item \textbf{Multimodal Low-rank Bilinear Attention (MLB)} \cite{an_5}: This model merges vision and language representation by Hadamard products. 
    \item \textbf{Multimodal Tucker Fusion (MUTAN)}\cite{an_6}:  This model joins vision and language in terms of a tensor decomposition. 
\end{itemize}

\begin{table}
  \caption{Ablation Study for losses}
  \centering
  \begin{tabular}{lllll}
    \toprule
    Approach   & Loss Ratio  & Q->A     & QA->R     & Q->AR \\
    \midrule
    Softmax & 1:4 & 58.1  & \textbf{62.18}   & 37.01  \\
    Softmax  & 1:1   & \bftab 63.76 & 61.61      & \textbf{39.76}  \\
    \midrule
    G-Softmax  & 1:4  & 59.32 & \textbf{61.07}      & 37.06  \\
    G-Softmax   & 1:1  & \bftab 64.54 & 61.05      & \bftab 40.28  \\
    \bottomrule
  \end{tabular}
  \label{loss_results}
\end{table}

\begin{table*}[ht]
  \begin{minipage}[b]{0.35\linewidth}
    \centering
    \includegraphics[width=60mm,height=40mm]{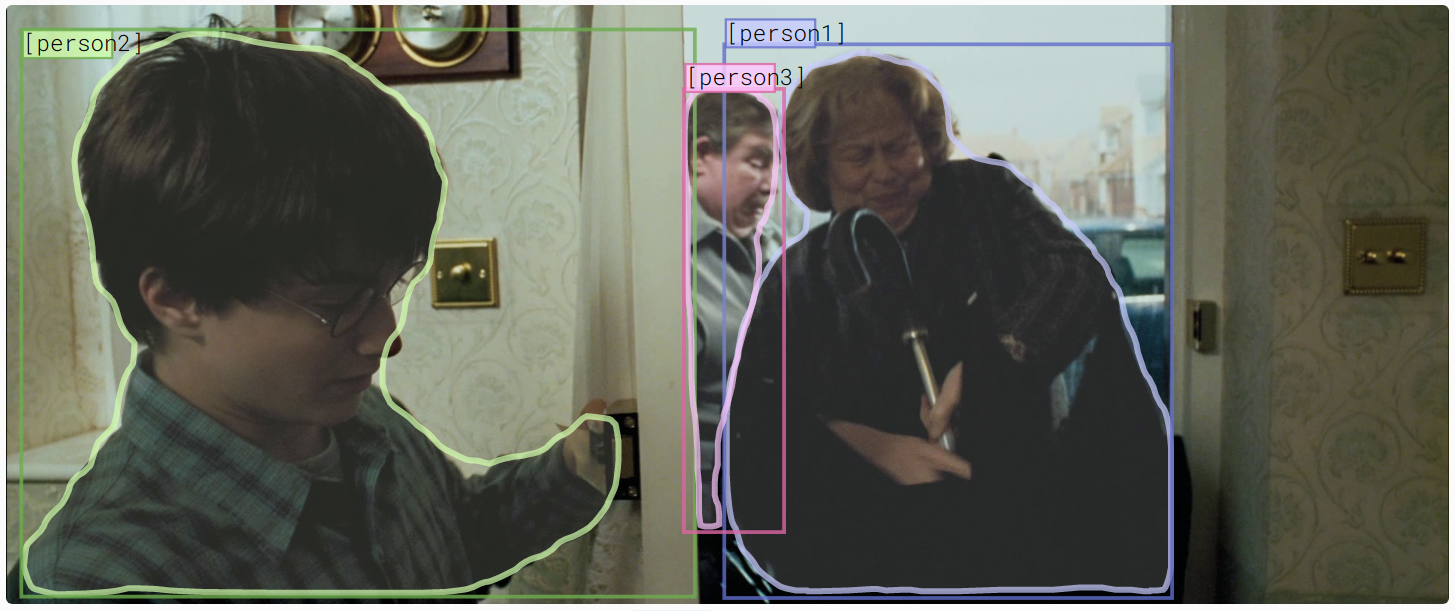}
    \label{fig:image}
  \end{minipage}%
  \begin{minipage}[b]{0.65\linewidth}
  \small
    \centering
    \begin{tabular}[b]{p{50mm}p{50mm}}
      \toprule
      \multicolumn{2}{p{100mm}}{\textbf{Question}:  How is [person1] feeling? } \\
      \midrule
      \textbf{Answers}  &  \textbf{Rationales} \\
      a) [person1] is feeling amused. & 
      a) [person1]'s mouth has wide eyes and an open mouth. \\
      b) \colorbox{green!50}{[person1] is upset and disgusted.} & 
      b) When people have their mouth back like that and their eyebrows lowered they are usually disgusted by what they see. \\
      c) [person1] is feeling very scared. & 
      c) [person3], [person2] and [person1] are seated at a dining table where food would be served to them. people unaccustomed to odd or foreign dishes may make disgusted looks at the thought of eating it. \\
      d) [person1] is is feeling uncomfortable with [person3]. & 
      d) \colorbox{green!50}{\parbox{45mm}{[person1]'s expression is twisted in disgust.}} \\
      \bottomrule
    \end{tabular}
  \end{minipage}
  \begin{minipage}[b]{0.35\linewidth}
    \centering
    \includegraphics[width=60mm,height=40mm]{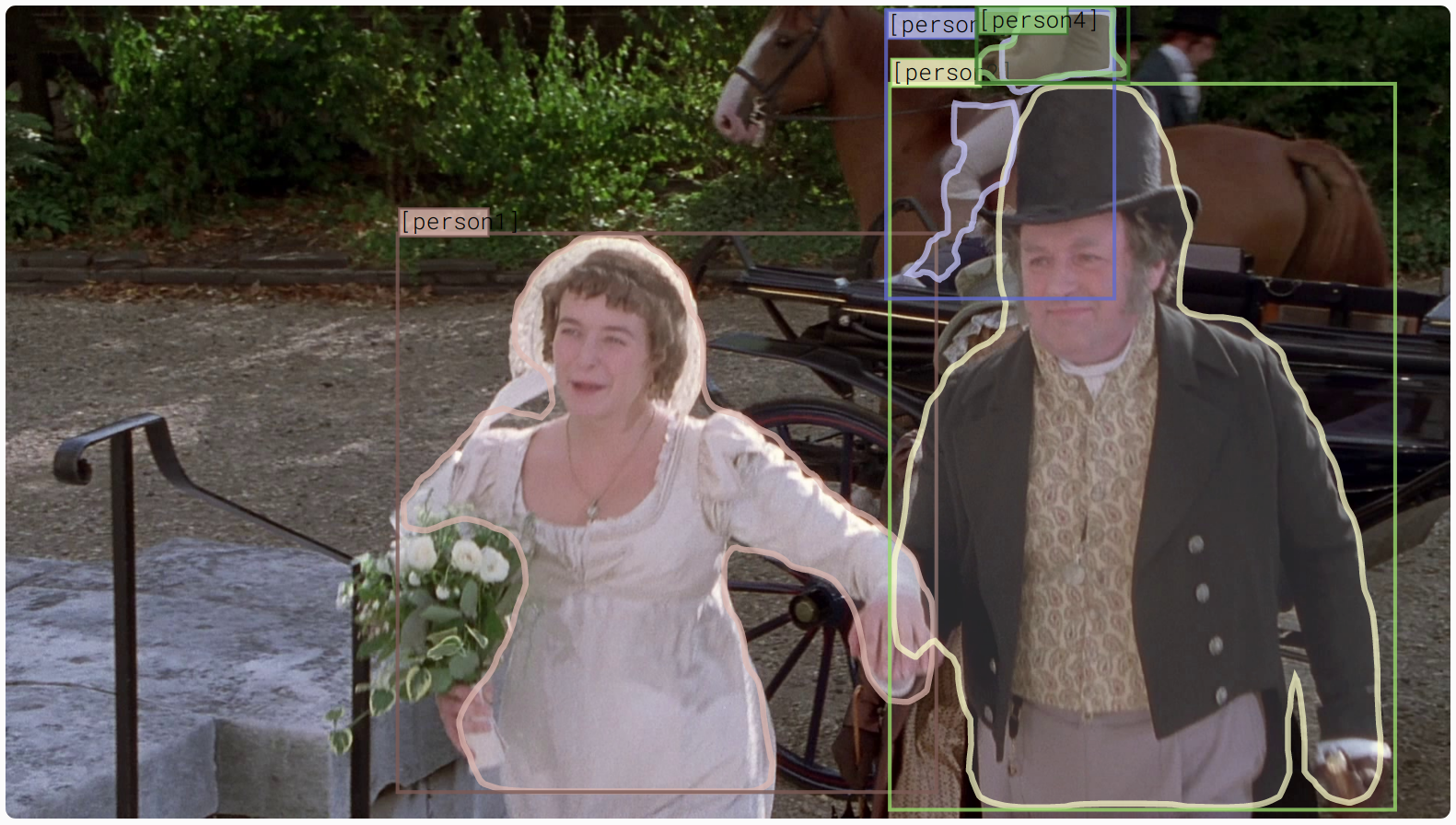}
    \label{fig:image}
  \end{minipage}%
  \begin{minipage}[b]{0.65\linewidth}
  \small
    \centering
    \begin{tabular}[b]{p{50mm}p{50mm}}
      \toprule
      \multicolumn{2}{p{100mm}}{\textbf{Question}:  Are [person1] and [person2] happy to get married? } \\
      \midrule
      \textbf{Answers}  &  \textbf{Rationales} \\
      a) Yes, [person1] and [person2] are in love. & 
      a) They're facing each other as they take their vows, while dressed in wedding attire. \\
      b) No [person1] and [person2] are not discussing something happy. &
      b) [person1] and [person2] are dressed formally, [person4] has on a wedding dress and there is draping above them. \\
      c) No, they are not. & 
      c) Both of them raise there arms up and slap hands together in a sign of celebration \\
      d) \colorbox{green!50}{Yes, they're both very happy today.} & 
      d) \colorbox{green!50}{\parbox{45mm}{They are both smiling and seem delighted.}} \\
      \bottomrule
    \end{tabular}
  \end{minipage}
  \begin{minipage}[b]{0.35\linewidth}
    \centering
    \includegraphics[width=60mm,height=40mm]{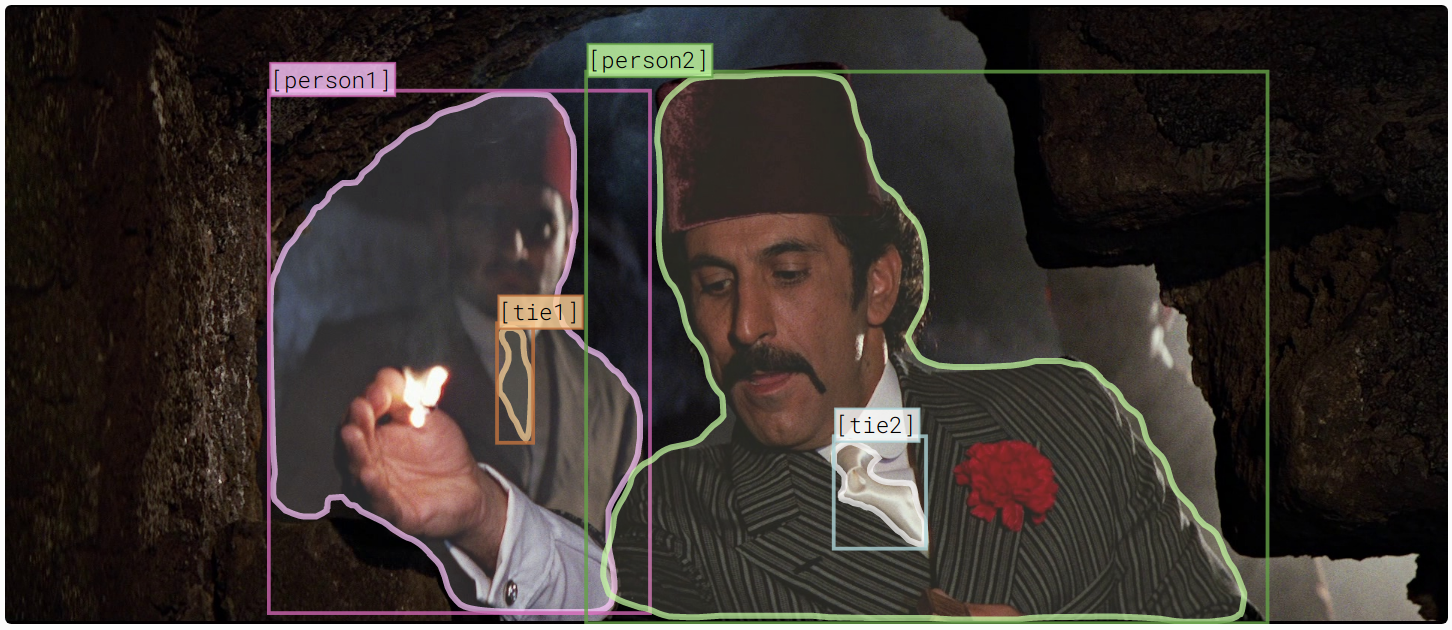}
    \label{fig:image}
  \end{minipage}%
  \begin{minipage}[b]{0.65\linewidth}
  \small
    \centering
    \begin{tabular}[b]{p{50mm}p{50mm}}
      \toprule
      \multicolumn{2}{p{100mm}}{\textbf{Question}:  What is [person2] doing? } \\
      \midrule
      \textbf{Answers}  &  \textbf{Rationales} \\
      a) Twirling on a dance floor & 
      a) \colorbox{cyan!50}{\parbox{45mm}{[person2] is looking down and examining something.}} \\
      b) Dealing cards to blackjack players. & 
      b) \colorbox{red!50}{[person1] has a serious expression.} \\
      c) \colorbox{green!50}{\parbox{45mm}{[person2] is contemplating something.}} & 
      c) People often consider the cards their opponent has in their hand when they want to win the game against them. \\
      d) [person3] is walking through some snow.  & 
      d) Sometimes when people stand with their hands on their hips and their eyes clothes it means that they are deep in thought.  \\
      \bottomrule
    \end{tabular}
  \end{minipage}
\caption{Qualitative Results: Examples of predictions made by our model. Green: When prediction matches the correct option. Blue: Correct option. Red: Wrong prediction.}
\label{r2c_qual}
\end{table*}



\subsection{Results}

\subsubsection{Model Performance Evaluation}

\textbf{Softmax}: Our Softmax approach performs well on Q->A task and acheives best result on QA->R task among the four approaches we tried. It is to be noted that for the QA->R task, we don't provide the correct answer as input to the model. Rather, a weighted average of answers (according to probabilities predicted by Q->A module) is provided to the rationale prediction module, unlike the baseline \cite{r2c}, which gives the correct answer as input.

\textbf{Gumbel-Softmax}: For gumbel-softmax, we anneal the temperature $\tau$ from 5 to 1 for 10 epochs and then keep it constant at 1. As can be seen from \ref{own_results}, this model gives the best result among all the approaches we used. Again, we provided the gumbel-softmax weighted average of answer representation to the rationale prediction module rather than the correct answer as was used in the baseline.

\textbf{Reinforcement Learning Sampling}: Surprisingly, the RL sampling based method performed poorly as compared to softmax and gumbel-softmax based method. The reason may be attributed to small number of samples being drawn for the expectation loss calculation. We were constrained by resource availability to limit the number of drawn samples to only 64 in each iteration. We leave this open to further exploration with a higher number of samples drawn in each training step.

\textbf{Direct Cross Entropy}: This method performs the worst. A plausible reason could be that the model fails to segregate subtle changes presented to it by the same rationale with different answers. We conclude that with the open-ended nature of rationales and answers and close similarity between them, the task becomes too difficult for the model, when asked to choose from sixteen possible choices.

\textbf{Comparison with new baseline}: We report the results of performance of our new baseline in table \ref{comp_new_baseline}. As can be seen, our best model (gumbel-softmax) performs strongly over the new baseline. It's better in all the three tasks - ~1\% better in Q->A task, ~3.5\% better in QA->R task and ~2\% better in Q->AR task. We conclude from this that gradient flow between the two Q->A and QA->R modules enabled by our end-to-end joint learning scheme, helps the network learn better answers for better/correct reasons.

\textbf{Comparison with State-of-the-art}: We provide comparison of our best model against state-of-the-art for visual common sense reasoning task. We also summarize results of other baselines reported in \cite{r2c}. As can be seen from table \ref{comp_results}, our Gumbel-softmax method performs better than the baseline \cite{r2c} in Q->A task. For the QA->R task, it is to be expected that our method should perform worse than the baseline as we are providing predicted answers, rather than the correct answer to the QA->R module. Still, our approach performs comparably against the baseline on the task. We conclude that our method learns to predict correct answers for the correct rationales.

\subsubsection{Ablation study for losses}

As we are using two losses - each for answer prediction and rationale prediction module, it makes sense to do an ablation study to study the effect of those losses (see figure \ref{r2c_loss}).

The task of rationale prediction is four times as difficult as answer prediction, since we are not providing correct answer to the rationale prediction module. As such, it makes sense to weight the QA->R loss four times as much as Q->A loss.
The figure shows it decreases the overall performance while improving the QA->R accuracy only slightly.

\section{Conclusion}

In this work, we aimed to enforce the cognitive learning in the newly formatted Visual Commonsense Reasoning task and  proposed an end-to-end trainable model
for joint learning of both answer and rationale. We explored four approaches to make the model differentiable,
 namely softmax, gumbel-softmax, cross entropy and reinforcement learning based sampling. These approaches enable the model to learn to make the correct predictions for the correct reasons without needing the ground-truth answers as input.
Although we are providing the predicted answer rather than the correct answer to the rationale prediction module as input (thus the performance is expected to decline), 
we show through experiments that our model is still able to perform competitively against the current state of the art. It even performed better than the state-of-the-art on Q->A task using gumbel-softmax. As state-of-the-art VQA models can perform at 75\% at best, we also introduced new baseline for VCR tasks which
feeds the correct answer 75\% time to the rationale prediction module.

We proposed one kind of method to condition the answer prediction on rationale prediction. However, our approach or the current state-of-the-art performs only at 44 \% at best, while the human accuracy is $\sim$ 90\%. There is a huge gap to be filled in to reach human level accuracy. One way to improve the overall performance of the system could be to imbibe domain/contextual knowledge separately to the model. Or it could be learned by the model on it's own through explorations using reinforcement learning. Another good future work
could be predicting the answer in the first step and then, use the predicted answer and question to generate reason. This generated reason could then be compared against the correct reason using an appropriate loss metric like BLUE score.

{\small
\bibliographystyle{ieee}
\bibliography{egpaper_final}
}

\end{document}